\documentclass[runningheads,sn-mathphys,Numbered]{sn-jnl}

\usepackage[ruled]{algorithm2e}
\usepackage{graphicx}%
\usepackage{multirow}%
\usepackage{amsmath,amssymb,amsfonts}%
\usepackage{amsthm}%
\usepackage{mathrsfs}%
\usepackage[title]{appendix}%
\usepackage{xcolor}%
\usepackage{textcomp}%
\usepackage{manyfoot}%
\usepackage{booktabs}%
\usepackage{algpseudocode}%
\usepackage{listings}%
\usepackage{fancyhdr}
\usepackage{float}

\theoremstyle{thmstyleone}%
\theoremstyle{thmstyletwo}%

\theoremstyle{thmstylethree}%

\begin{document}

\title[Article Title]{BdSpell: A YOLO-based Real-time Finger Spelling System for Bangla Sign Language}


\author*[1,2]{\fnm{Naimul} \sur{Haque}}\email{naimul011@gmail.com}

\author[2,3]{\fnm{Meraj} \sur{Serker}}\email{meraj.serker25@gmail.com}
\equalcont{These authors contributed equally to this work.}

\author[1,2]{\fnm{Tariq} \sur{Bin Bashar}}\email{tariqbinbashar3203@gmail.com}
\equalcont{These authors contributed equally to this work.}

\affil*[1]{\orgdiv{Computer Science and Engineering}, \orgname{Uttara University}}

\affil[2]{\orgdiv{Computer Science and Engineering}, \orgname{Daffodil International University}}

\affil[3]{\orgdiv{Computer Science and Engineering}, \orgname{Manarat International University}}


\abstract{In the domain of Bangla Sign Language (BdSL) interpretation, prior approaches often imposed a burden on users, requiring them to spell words without hidden characters, which were subsequently corrected using Bangla grammar rules due to the missing classes in BdSL36 dataset. However, this method posed a challenge in accurately guessing the incorrect spelling of words. To address this limitation, we propose a novel real-time finger spelling system based on the YOLOv5 architecture. Our system employs specified rules and numerical classes as triggers to efficiently generate hidden and compound characters, eliminating the necessity for additional classes and significantly enhancing user convenience. Notably, our approach achieves character spelling in an impressive 1.32 seconds with a remarkable accuracy rate of 98\%. Furthermore, our YOLOv5 model, trained on 9147 images, demonstrates an exceptional mean Average Precision (mAP) of 96.4\%. These advancements represent a substantial progression in augmenting BdSL interpretation, promising increased inclusivity and accessibility for the linguistic minority. This innovative framework, characterized by compatibility with existing YOLO versions, stands as a transformative milestone in enhancing communication modalities and linguistic equity within the Bangla Sign Language community.

}




\maketitle

\section{Introduction}\label{sec1}

In a world increasingly connected through technology, accessibility, and inclusivity are of paramount importance. The Real-Time Bangla Finger Spelling for Sign Language project represents a significant endeavor to bridge communication gaps for individuals with hearing and speech impairments in Bangladesh. This pioneering project aims to develop an advanced computer vision system capable of accurately detecting and interpreting Bangla finger spelling gestures in real-time, empowering users to communicate effectively using sign language.

The project leverages the power of YOLOv5 \cite{YOLOv5}, a cutting-edge object detection algorithm renowned for its speed and precision. By harnessing YOLOv5’s capabilities, the system seeks to enable real-time recognition of Bangla finger spelling gestures for digits and alphabets, paving the way for seamless communication and meaningful interactions for individuals with impairments.

With a primary objective of creating a highly accurate and efficient computer vision model, the project places a strong emphasis on developing a robust dataset encompassing various hand orientations, lighting conditions, and backgrounds. This diversity ensures the system’s adaptability to real-world scenarios and enhances its ability to accurately interpret Bangla finger spelling gestures. By evaluating the system’s performance based on metrics such as Mean Average Precision, Precision, and Recall, the project ensures the model’s reliability and effectiveness in avoiding false positives and negatives during real-time inference.

Furthermore, prior approaches \cite{grammer} in the domain of Bangla Sign Language (BdSL) interpretation often imposed a burden on users, requiring them to spell words with missing characters, which were subsequently corrected using Bangla grammar rules due to the missing classes in BdSL36 dataset. However, this method posed a challenge in accurately guessing the incorrect spelling of words. To address this limitation, we propose a novel real-time finger spelling system based on the YOLOv5 architecture. Our system employs specified rules and numerical classes as triggers to efficiently generate hidden and compound characters, eliminating the necessity for additional classes and significantly enhancing user convenience. Another YOLOv4 \cite{wordgeneration} based system was proposed which takes 60 sec for each character detection using a total of 49 different classes. Our approach achieves character spelling in an impressive 1.32 seconds with a remarkable accuracy rate of 98\%. Furthermore, our YOLOv5 model, trained on 9147 images, demonstrates an exceptional mean Average Precision (mAP) of 96.4\%.

\begin{itemize}
    \item Our YOLOv5 model, trained on 9147 images, demonstrates an exceptional Mean Average Precision (mAP) of 96.4\%, showcasing its impressive accuracy.
    
    \item Our proposed system eliminates the need for additional classes, significantly reducing the computational cost while enhancing user convenience in comparison to previous approaches.
    
    \item Our approach achieves character spelling in an impressive 1.32 seconds with a remarkable accuracy rate of 98\%, representing a substantial speed improvement over previous systems.
    
    \item We implemented a technique involving thresholding the mean running cumulative of the confidence of the detection for spelling characters, further enhancing the accuracy and reliability of our system.
\end{itemize}

This paper will delve into the project’s methodology, outline the implementation steps, and discuss the dataset preparation, model training, and evaluation processes. Additionally, it will highlight the potential impact of the Real-Time Bangla Finger Spelling for Sign Language project in fostering inclusivity and empowering individuals with hearing and speech impairments to communicate effectively with others. Through this research endeavor, we aspire to contribute to building a more accessible and inclusive society that values effective communication for all, reaffirming the significance of technology in promoting a more connected and empathetic world.
\section{Related Works}\label{sec2}

Nanda et al. \cite{yolov3} explored real-time sign language conversion using the YOLOv3 algorithm. They focused on American Sign Language (ASL) and employed default hyperparameters for training. In \cite{thaiyolo}, researchers tackled Thai Sign Language classification using the YOLO algorithm. Their dataset comprised 25 signs with 15,000 instances for training. They achieved an mAP of 82.06\% in complex background scenarios. Arabic Sign Language Recognition and Speech Generation were studied in \cite{arabicsl} using Convolution Neural Networks. The authors integrated Google Translator API for hand sign-to-letter translation and gTTs for speech generation.

For Bangla Sign Language (BdSL) detection, \cite{siftncnn} employed a method involving RGB-to-HSV color space conversion, followed by feature extraction using the Scale Invariant Feature Transform (SIFT) Algorithm for 38 Bangla signs. A simple neural network was used for Bangla alphabet classification in \cite{freeman}, employing the YCvCr color map for input images. They used the Canny edge detector and Freeman Chain Code for feature extraction. A real-time Bangladeshi Sign Language Detection method using Faster R-CNN was proposed in \cite{rcnn}, achieving an accuracy of 98.2\% and a detection time of 90.03 milliseconds, using a dataset containing 10 different sign letter labels.

Hossen et. al, \cite{dcnnbdsl} presented a Deep Convolutional Neural Network-based method for Bengali Sign Language Detection, utilizing a diverse dataset of 37 signs with various backgrounds and skin colors. In \cite{recognitionbdslcnn}, a dataset comprising 7052 samples of 10 numerals and 23864 samples of 35 characters of Bangla Sign Language was introduced. The Convolutional Neural Network was employed for accurate classification. Sarker et. al, \cite{speech} proposed a Bangla Sign language-to-speech generation system using smart gloves, sensors, and a microcontroller. They employed Levenshtein distance for word matching in the database for sign recognition. In the domain of finger-spelling, Li et al. \cite{recognition} presented a real-time finger-spelling recognition system using a convolutional neural network (CNN) architecture. They achieved high accuracy in recognizing finger-spelling gestures in real-time scenarios.

Another research \cite{grammer} used Unveiling an Innovative Algorithm for Accurate Hand-Sign-Spelled Bangla Language Modelling. Authors used Bangla grammatical rules to get hidden characters. And they generated independent vowels using those rules. Dipon et al, \cite{wordgeneration} used YOLOv4 as the object detection model for Real-Time Bangla Sign Language Detection with Sentence and Speech Generation. Detection time of 60 seconds for every word. Rafiq et. al, proposed a real-time vision-based Bangla sign language detection system \cite{visionyolo} using the YOLO algorithm. They used a dataset consisting of 10 Bangla signs and achieved an mAP of 92.5\%. Talukder et. al, proposed a real-time Bangla sign language detection system using the YOLOv4 \cite{49classesyolov4}object detection model. They used a dataset consisting of 49 different classes, including 39 Bangla alphabets, 10 Bangla digits, and three new proposed signs. They achieved an mAP of 95.6.																			

These finger spelling recognition papers complement the project's focus on Real-Time Bangla Finger Spelling for Sign Language, contributing valuable insights and methodologies to the broader domain of sign language recognition and communication accessibility.

\section{Dataset}\label{sec3}
We utilized the BdSL36 dataset \cite{bdsl36}, purposefully curated for Bangladesh Sign Language recognition systems by Oishee Bintey Hoque et al. This dataset underwent meticulous preparation across five stages to ensure robustness and versatility.

The process commenced with Image Collection, conducted through extensive research at a deaf school to identify 36 practical Bangla sign letters used in daily communication. Ten volunteers captured raw images using phone cameras or webcams. Subsequently, BdSL experts individually assessed and filtered the images to retain those aligning with the appropriate sign style. This curation yielded 1200 images across the classes.

\begin{figure}[h!]%
\centering
\includegraphics[width=0.9\textwidth]{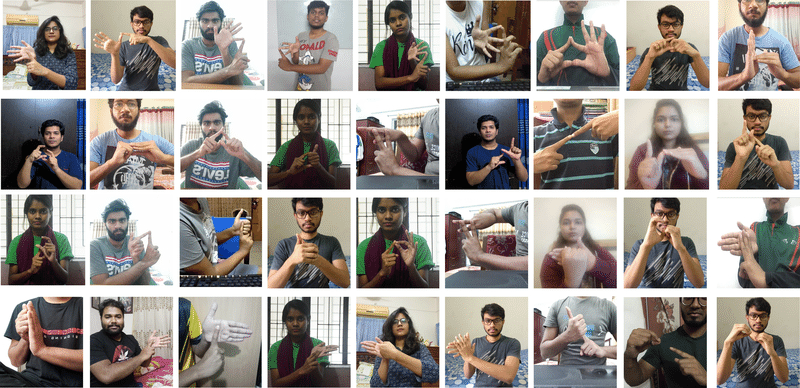}
\caption{ Example images of initially collected BdSL36 dataset. Each image represents a different BdSL sign letter. Images are serially organized according to their
class label from left to right. }\label{bdsl36}
\end{figure}

Raw Image-Data Augmentation addressed the need for accurate sign letter detection under varying conditions. Manual augmentation techniques, encompassing affine and perspective transformations, contrast adjustments, noise addition, cropping, blurring, rotation, and more, were applied, resulting in the BdSL36v1 dataset containing approximately 26,713 images with an average of 700 images per class. Few sample images of the dataset are shown in the Figure \ref{bdsl36}.

Dataset preparation involved a comprehensive and adaptable approach to BdSL recognition. The meticulous stages of image collection, augmented data generation, and background augmentation ensure that the BdSL36 dataset authentically captures real-world scenarios, rendering it a valuable resource for advancing the field of sign language recognition and detection. Additionally, we further annotated 9,187 BdSL36 dataset images and split it into 6,427 training sets, 1,828 validation sets, and 932 testing sets.

\section{Bangla Alphabets}\label{sec3}
Bengali is written from left to right, like the majority of other languages, and there are no capital characters. The letters have a continuous line at the top, and there are conjuncts, upstrokes, and downstrokes in the script. There are a total of 60 characters including 11 vowels (Sworoborna), 39 consonants (Byanjanbarna), and 10 numerals. 

\begin{figure}[h!]%
\centering
\includegraphics[width=0.9\textwidth]{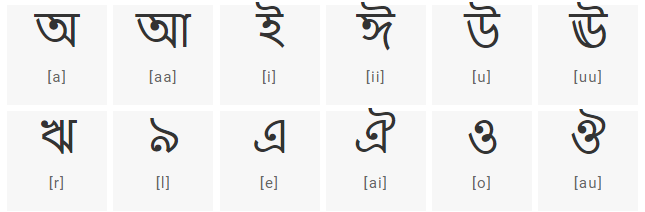}
\caption{Bangla Vowels: A Visual Representation of the Complete Set of Vowels in the Bengali Alphabet}\label{vowels}
\end{figure}

\begin{figure}[h!]%
\centering
\includegraphics[width=0.9\textwidth]{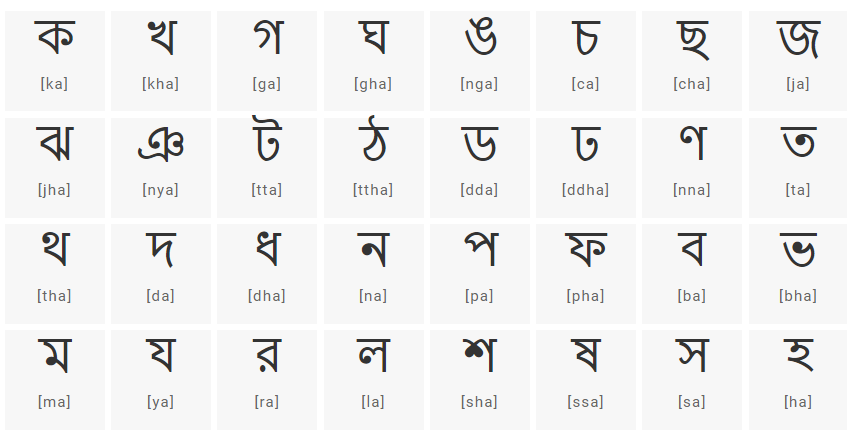}
\caption{Bangla Consonants: Illustrating the Entire Array of Consonant Characters in the Bengali Alphabet.}\label{consonets}
\end{figure}

\begin{figure}[h!]%
\centering
\includegraphics[width=0.9\textwidth]{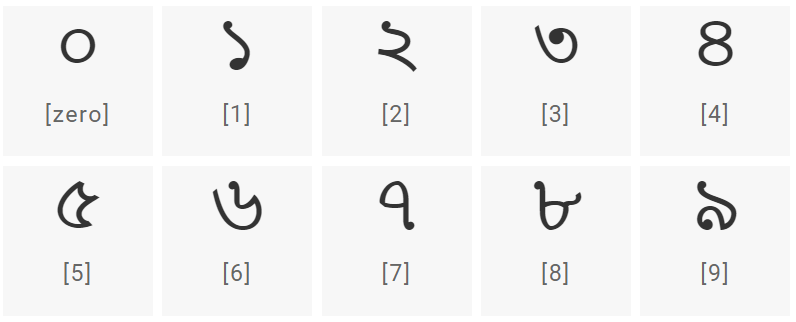}
\caption{Bangla Numerals: Displaying the Full Range of Numerical Digits in the Bengali Script.}\label{numerals}
\end{figure}

\begin{figure}[h!]%
\centering
\includegraphics[width=0.9\textwidth]{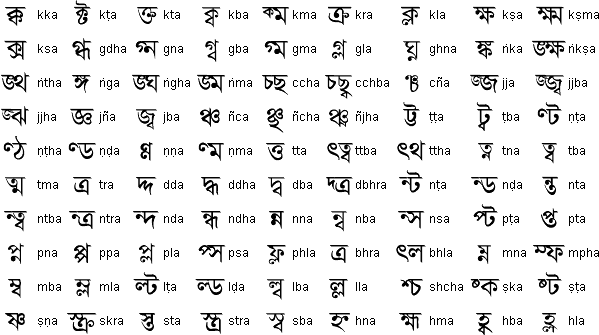}
\caption{Bangla Compound Characters: Essential Combinations in the Bengali Alphabet. Bangla Compound Characters, known as "Yuktakshar" in Bengali, are formed by combining two or more basic characters from the script. These combinations create unique characters that represent specific phonetic sounds not present in the basic alphabet. They are pivotal in accurately transcribing words and phrases in the Bengali language. This figure showcases some of the most commonly used compound characters in the Bengali script, highlighting their significance in phonetic representation}\label{compound}
\end{figure}

\begin{figure}[h!]%
\centering
\includegraphics[width=0.9\textwidth]{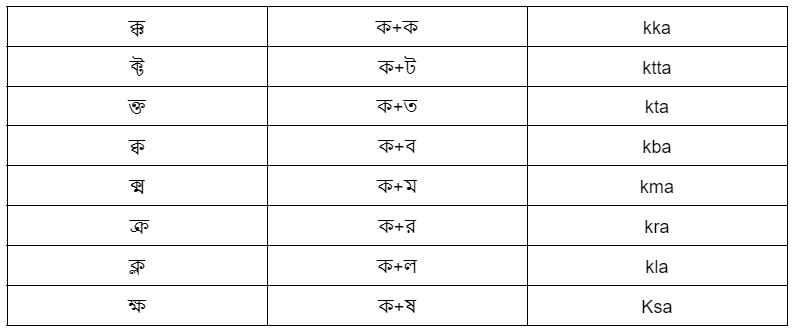}
\caption{An illustration of consonant combinations resulting in compound characters in the Bengali script. The first column displays select compound characters, while the second column demonstrates the amalgamation of consonant characters that give rise to these combinations, showcasing the script's phonetic intricacies.}\label{compound_formation}
\end{figure}

Sworoborna, shown in Figure \ref{vowels} refers to those letters in Bengali that can be spoken on their own. These are the characters that represent individual vowel sounds and can be pronounced independently without the need for a consonant. These vowel characters are an integral part of the Bengali script and play a crucial role in forming words and conveying meaning. They are combined with consonant characters (Byanjanbarna) to create syllables, which ultimately from words. 
In the Bengali language, consonants are known as “Byanjonborno”, shown in Figure \ref{consonets}. These are letters that cannot be pronounced on their own and need to be combined with vowels to create a complete sound. Bengali has a total of 32 consonant letters, which play a crucial role in forming meaningful words and expressions. When a consonant is combined with a vowel, it forms a syllable, which is the fundamental unit of pronunciation in Bengali. This combination of consonants and vowels allows speakers to articulate a wide range of sounds and convey various meanings.
Bengali has its own set of numeric symbols to represent numbers and fractions, shown in Figure \ref{numerals}. These symbols are used for numerical representation in various contexts, such as writing numbers, expressing quantities, and indicating fractions. 
In Bengali, compound characters are formed by combining two or more consonants to create a single character. Some of the most commonly used compound characters are shown in Figure \ref{compound}. In Bengali, compound characters are formed by combining two or more consonants to create a single character. Some of the most commonly used compound characters are shown in Figure \ref{compound}. The formation of these compound characters has been illustrated using some examples in Figure \ref{compound_formation}.

\section{Objection Detection Evaluation Metric}\label{sec3}
In object detection tasks, evaluating the performance of a model is crucial to understanding its accuracy and effectiveness. One of the most commonly used evaluation metrics for object detection is Mean Average Precision (mAP). In this section, we will elaborate on how mAP works and the different components involved in its calculation.

\subsection{From Prediction Score to Class Label}

In object detection, the model predicts bounding boxes around objects along with their corresponding class labels and confidence scores. The prediction score represents how confident the model is in its prediction. To convert this prediction score into a class label, a threshold is applied. If the confidence score for a prediction exceeds the threshold, it is classified as a positive detection with the associated class label; otherwise, it is considered a negative detection.

$\text{Class Label} = \begin{cases} 1 & \text{if } \text{Confidence Score} > \text{Threshold} \\ 0 & \text{otherwise} \end{cases}$

\subsection{Detection Performance Metrices}

Precision is a metric that measures the accuracy of the model's predictions for a particular class. It is defined as the ratio of true positive (TP) detections to the sum of true positive and false positive (FP) detections:\\\\
$Precision (P) = \frac{TP}{TP + FP}$.\\\\ Recall, also known as True Positive Rate (TPR) or Sensitivity, measures the model's ability to find all the positive instances for a particular class. It is defined as the ratio of true positive detections to the sum of true positive and false negative (FN) detections:\\\\ $Recall (R) = \frac{TP}{TP + FN}$\\\\ The Precision-Recall (PR) curve is a graphical representation of the precision and recall values at various confidence score thresholds. It helps to visualize how precision and recall change as we vary the confidence threshold for positive detections. The PR curve is obtained by plotting precision on the y-axis and recall on the x-axis.

Average Precision (AP) is a single scalar value that summarizes the Precision-Recall curve for a specific class. It is calculated by computing the area under the Precision-Recall curve. The mathematical formula for AP is as follows:\\\\$AP = \int_{0}^{1} \text{precision}(r) \, dr$\\\\;where 
$precision(r)$ is the precision at a given recall value $r$ in the Precision-Recall curve.

\subsection{Intersection over Union (IoU)}

$IoU$ is a critical concept in object detection evaluation. It measures the overlap between the predicted bounding box and the ground truth bounding box. IoU is computed as the ratio of the area of intersection between the two bounding boxes to the area of their union:

$IoU = \frac{\text{Area of Intersection}}{\text{Area of Union}}$

\begin{figure}[h!]%
\centering
\includegraphics[width=0.5\textwidth]{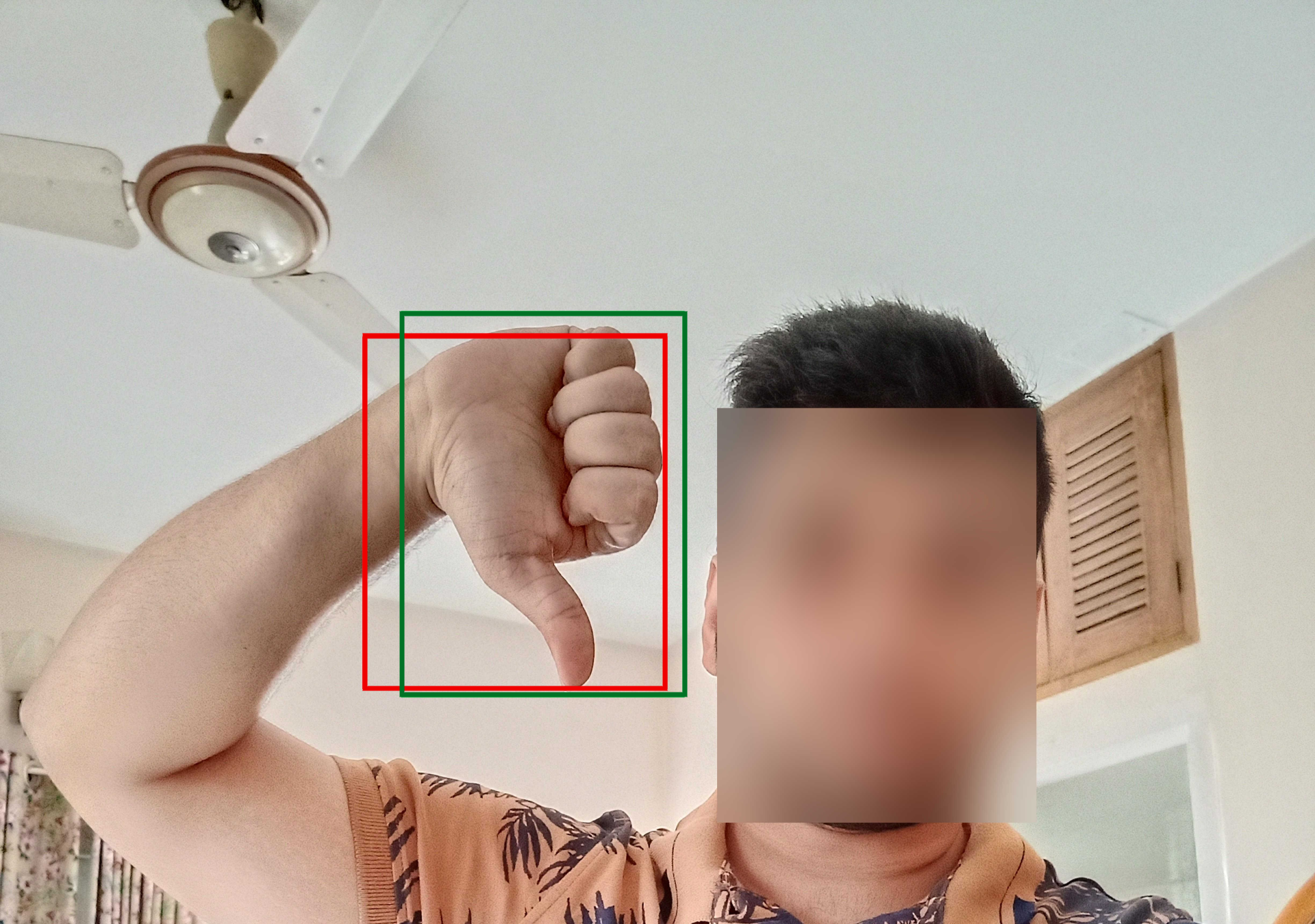}
\caption{ In this illustrative image, the concept of IoU is demonstrated. The predicted bounding box (in red) and the ground truth bounding box (also in red) showcase the overlapping area used to compute IoU. A higher IoU value signifies accurate object localization, while a lower value indicates less precise detection.}\label{iou}
\end{figure}

The example shown in Figure \ref{iou} shows how IoU is calculated.

\subsection{Mean Average Precision (mAP)}
$mAP$ is the average of AP values calculated for all the classes in the dataset. It provides a comprehensive evaluation of the model's overall performance across different classes and confidence score thresholds. The mathematical formula for $mAP$ is as follows:\\\\$mAP = \frac{1}{N} \sum_{i=1}^{N} AP_i$\\\\Where $N$ is the number of classes and $AP_i$ is the Average Precision for $class_i$.

By using these evaluation metrics and mathematical formulas, we can effectively assess the performance of an object detection model, identify areas of improvement, and fine-tune the model to achieve better accuracy and reliability in detecting objects of interest.
\section{Methodology}\label{sec3}
To develop a YOLO-based real-time finger spelling model for the BDSL36 dataset, which encompasses 36 recognized characters along with several derived characters, we will follow a systematic methodology. The BDSL36 dataset contains a diverse set of characters represented by Unicode values, each associated with a corresponding finger spelling label. To devise a comprehensive system capable of recognizing hidden or derived characters, we introduced a set of key components within our methodology. These components include \textbf{Recognized Character Detection}, \textbf{Independent Vowel Transformation}, \textbf{Hidden Character Generation}, and \textbf{Trigger Handling}, each playing a vital role in the fingerspelling process.

\begin{figure}[h!]%
\centering
\includegraphics[width=\textwidth]{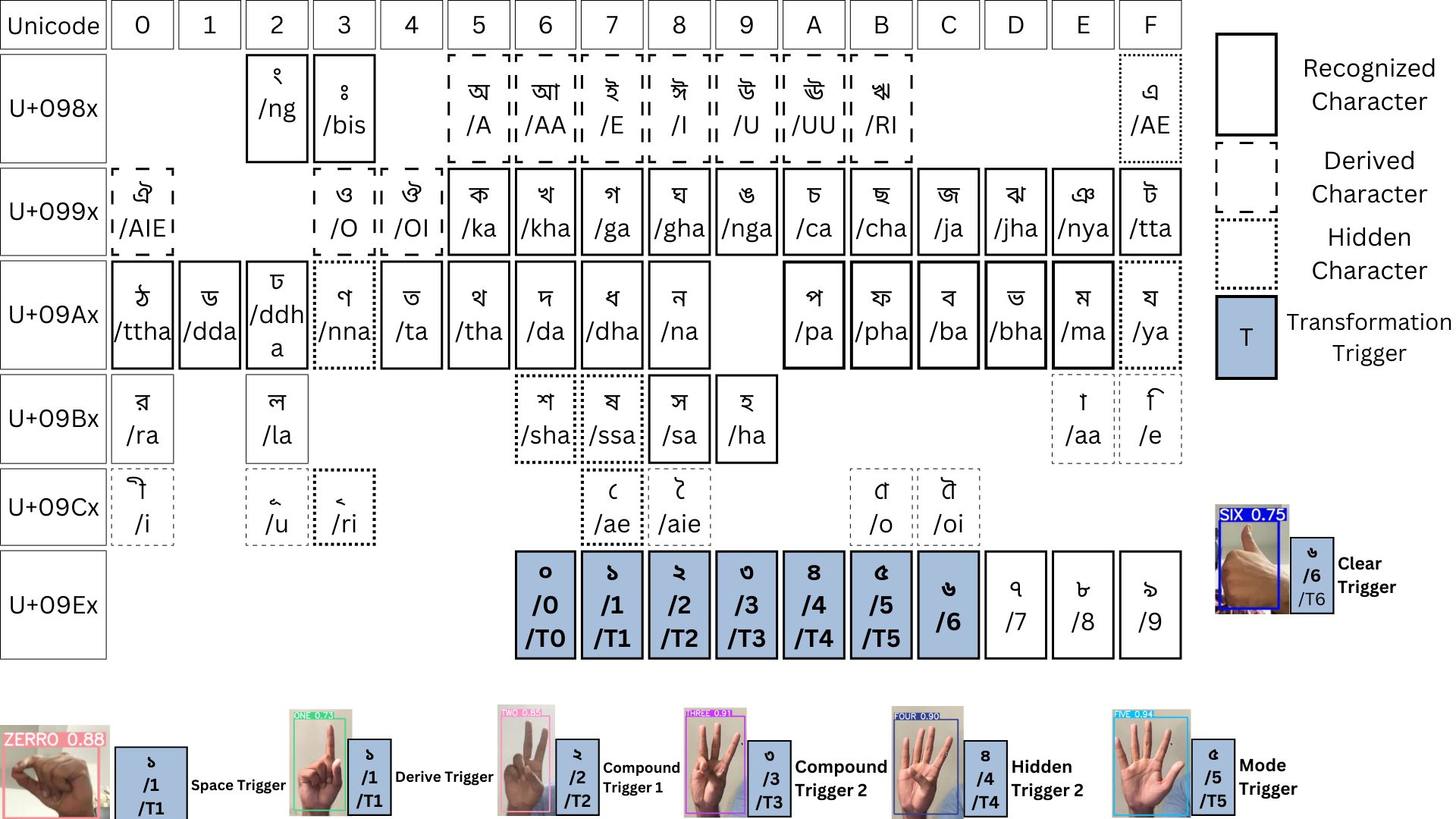}
\caption{This image showcases a comprehensive array of Bangla characters, thoughtfully organized into four distinct categories: recognized characters, derived characters, hidden characters, and Trigger characters. The top-right corner provides a visual taxonomy for easy reference. At the bottom, a specific example of a Trigger character is thoughtfully presented, offering a practical illustration of this unique character type in the Bangla script.}\label{bengali_alphabets}
\end{figure}

\begin{enumerate}
\item \textbf{Recognized Character Detection}: Recognized characters in our finger spelling recognition system are identified through the use of confidence scores $c_{i}(t)$ generated by the YOLOv5 model at time $t$ for the detection class $i$. All the available recognized characters are shown in Figure \ref{bengali_alphabets}. To qualify as a recognized character, the cumulative running mean of confidence must surpass a specified threshold $\delta$, as determined by the formula for the running cumulative confidence:\\

$\sum_{t}^{T} \frac{1}{N} \cdot \sum_{i}^{N} c_i(t) > \delta$\\

The $N$ is the number of detections of class $i$ at time frame $t$. This threshold $\delta$ ensures that only characters with consistently high confidence scores are selected. The recognized characters originate from the BDSL36 \cite{bdsl36} dataset, having undergone extensive training for object detection. They serve as the cornerstone for identifying both overt and derived characters within our system, providing a robust foundation for accurate recognition.

\item \textbf{Independent Vowel Transformation}: In written Bangla, it's important to note that there are two sets of vowels: independent and dependent. Dependent vowels are essentially the same as independent vowels, but they can only be written after consonants in the Bangla language, hence the name "dependent vowels". Since our system utilizes only one set of vowels because of the limitation of the BdSL36 dataset, we assume by default that a recognized vowel is a dependent vowel. This assumption stems from the fact that dependent vowels are more frequently used in Bangla spelling due to their characteristic placement after consonants. This nuanced understanding enables our model to accurately transcribe and recognize vowels in the Bengali script. Independent vowels in our system are not directly recognized. They are derived from the dependent vowels, which are recognized by the model. These recognized vowels transition into independent vowels when trigger characters are detected following a recognized character or derived character. This distinction allows us to handle the recognition of vowels effectively. The sets of vowels and a simple demonstration of the transformation that has been shown in Figure \ref{transformation}.
  
\item \textbf{Hidden Character Generation}: Hidden characters are a unique aspect of our system. These characters are not provided in the BDSL36 dataset except for the independent variables. They play a pivotal role in accurately representing the Bengali script. These hidden characters, though not as widely recognized, are essential for spelling numerous Bangla words. We carefully define and create these hidden characters, ensuring they are not directly recognizable by the model. This distinctive feature empowers our system to bridge the gap between the limitations of existing datasets and the comprehensive representation of the Bengali language. Using the provided rules stated in Figure \ref{transformation}.

\item \textbf{Trigger Handling}: Triggers are specific characters ranging from T0 to T7, which are shown in the Figure \ref{bengali_alphabets}. They operate
exclusively in the Textual mode of our finger spelling system. The specific functions and roles of these trigger characters will be discussed in detail later in our
methodology. They serve as key elements for recognizing derived characters and their dependencies. Detecting and handling these Trigger characters is a crucial part of our finger-spelling methodology.  
.

\end{enumerate}
	
By incorporating these components into our methodology, we ensure a holistic approach to finger-spelling recognition. Our system not only recognizes the characters available in the BDSL36 dataset but also effectively handles hidden characters, transitions dependent vowels into independent vowels when triggers are detected, and utilizes trigger characters for recognizing derived characters. This comprehensive approach enables us to create a robust and versatile finger spelling recognition system that accommodates the complexities of the Bengali language, where hidden and derived characteristics play significant roles in communication.

\subsection{BdSL Finger-Spelling}

To derive characters that are not present in the BDSL36 dataset, we can establish transformation rules based on the provided mappings in Figure \ref{transformation}. There are four types of characters that need to be derived: Independent Vowels, Hidden Characters, Compound Characters with two characters, and Compound Characters using three characters. 

\subsubsection{Single Character Transformation}
\begin{figure}[h!]%
\centering
\includegraphics[width=\textwidth]{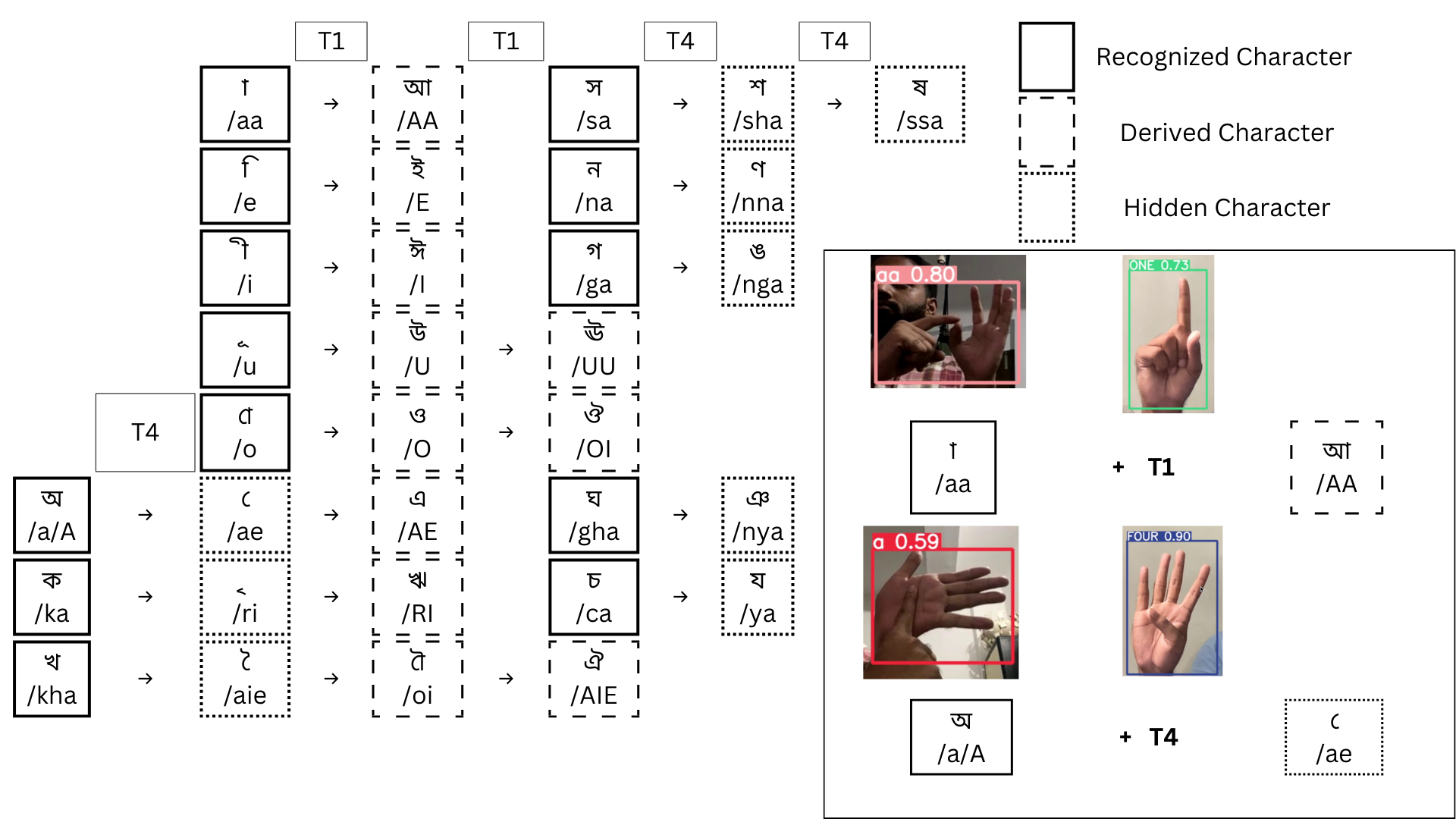}
\caption{In this visual representation, we observe the Transformation of Dependent Vowels to Independent when Trigger T1 is finger-spelled while hidden characters are derived from other characters using T4. In the lower right-hand corner of this image, we are presented with two illustrative examples. In the first one, we witness the transformation of the character "/aa" into "/AA" as a result of the influence of 'Trigger T1.' and in the second one, we can observe the transformation of the sequence "/a/A" into the character "/ae." This transformation is facilitated by the operation of 'T4.'}
\label{transformation}
\end{figure}

According to the provided figure \ref{transformation}, when someone finger spells any recognized dependent vowel character and then follows it with Trigger 1 (T1) finger-spelling, the dependent vowel character will be replaced with the corresponding independent vowel character. Additionally, hidden characters are derived from recognized characters when Trigger 4 (T4) is finger-spelled according to the figure \ref{transformation}.

\subsubsection{Compound Character Derivation}
\begin{figure}[h!]%
\centering
\includegraphics[width=0.9\textwidth]{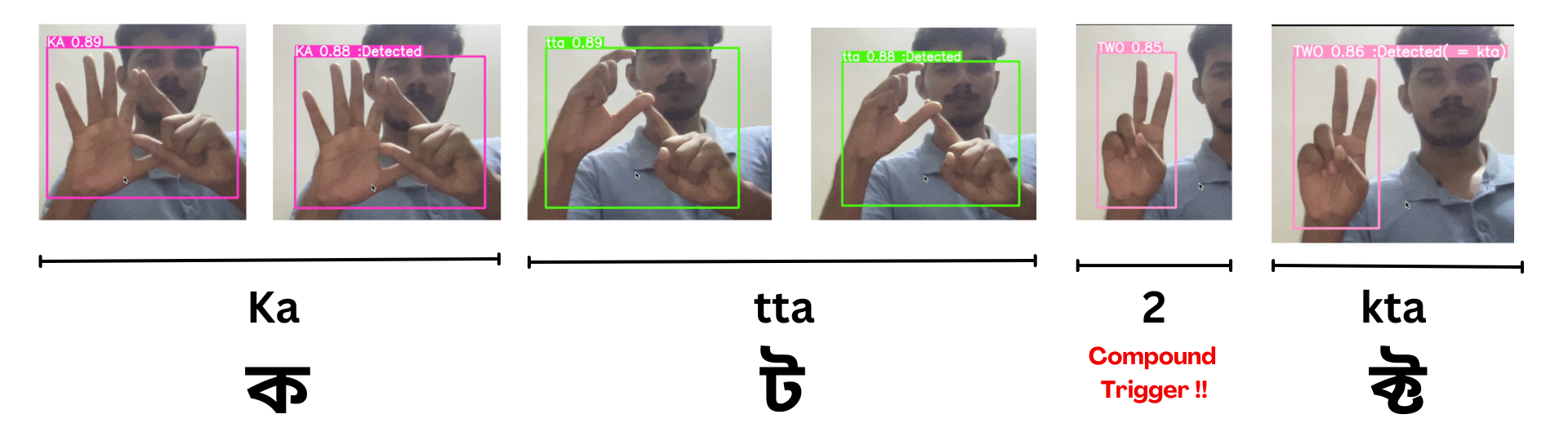}
\caption{Formation of Compound Character 'kta' by Combining 'ka' and 'tta' in Bengali Script. The illustration shows how a hand signer finger spell compound character after detecting the Trigger T2 which automatically remove the last two characters and append their corresponding compound character in our spelled word.}\label{compounding}
\end{figure}

\begin{figure}[h!]%
\centering
\includegraphics[width=\textwidth]{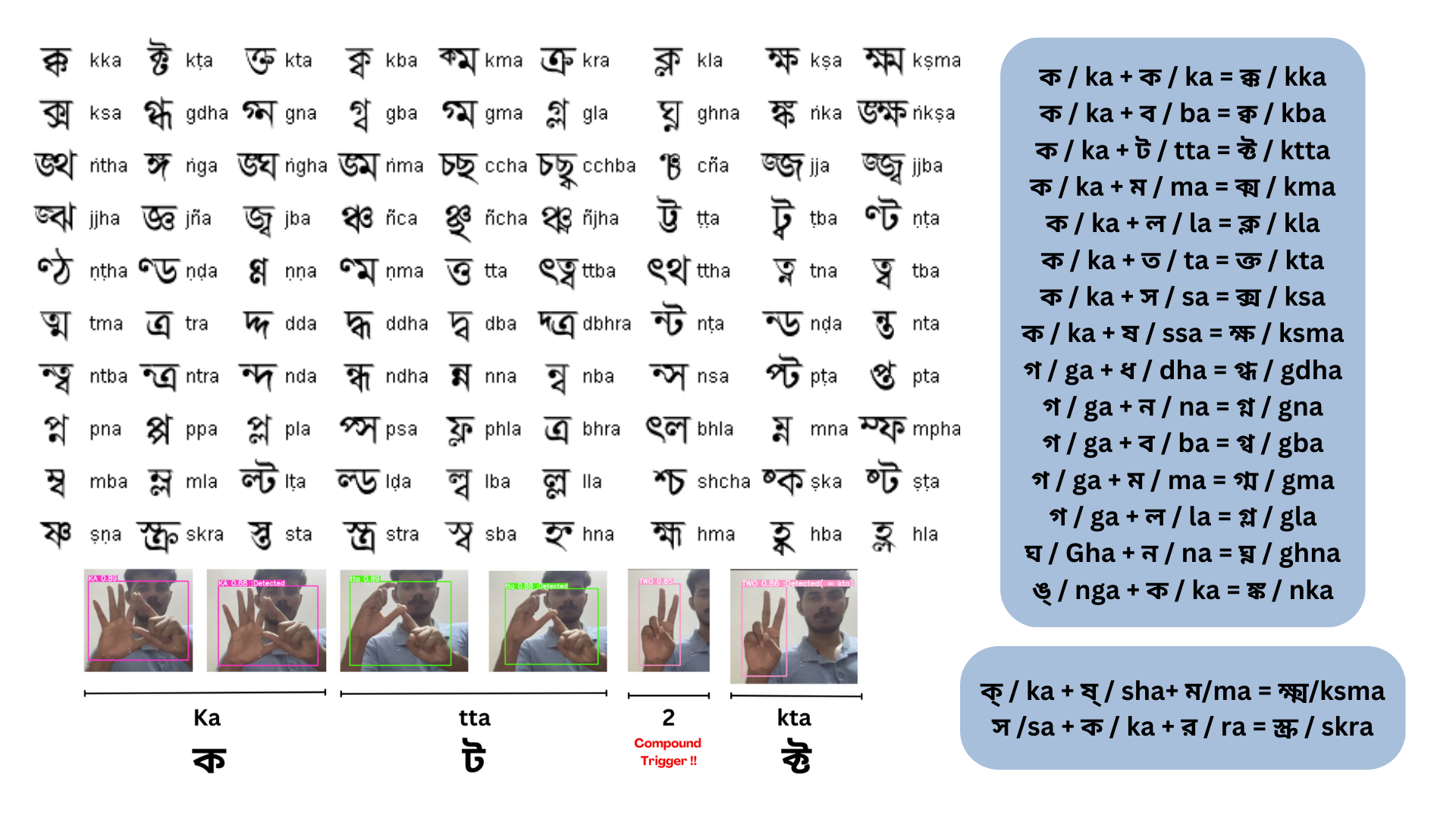}
\caption{In this visual representation, We are presented with a comprehensive overview of Bangla finger spelling. On the left side of the image, we can observe a visual representation of the compound characters, where two or more individual characters combine to form a unique and meaningful sign. the right side of the picture provides a detailed set of rules and guidelines. These rules outline the correct formation and usage of compound characters, both for two-character and three-character combinations.}\label{compound_derivation}
\end{figure}

In this subsection, we explore the process of deriving compound characters from recognized characters using triggers T2 and T3, along with the Bangla compound character dictionary. These triggers allow us to create compound characters by combining individual characters. The dictionary and the derivation process are illustrated in Figure \ref{compound_derivation}. The Figure \ref{compounding} shows how to spell a compound character in real time 

\begin{figure}[h!]%
\centering
\includegraphics[width=\textwidth]{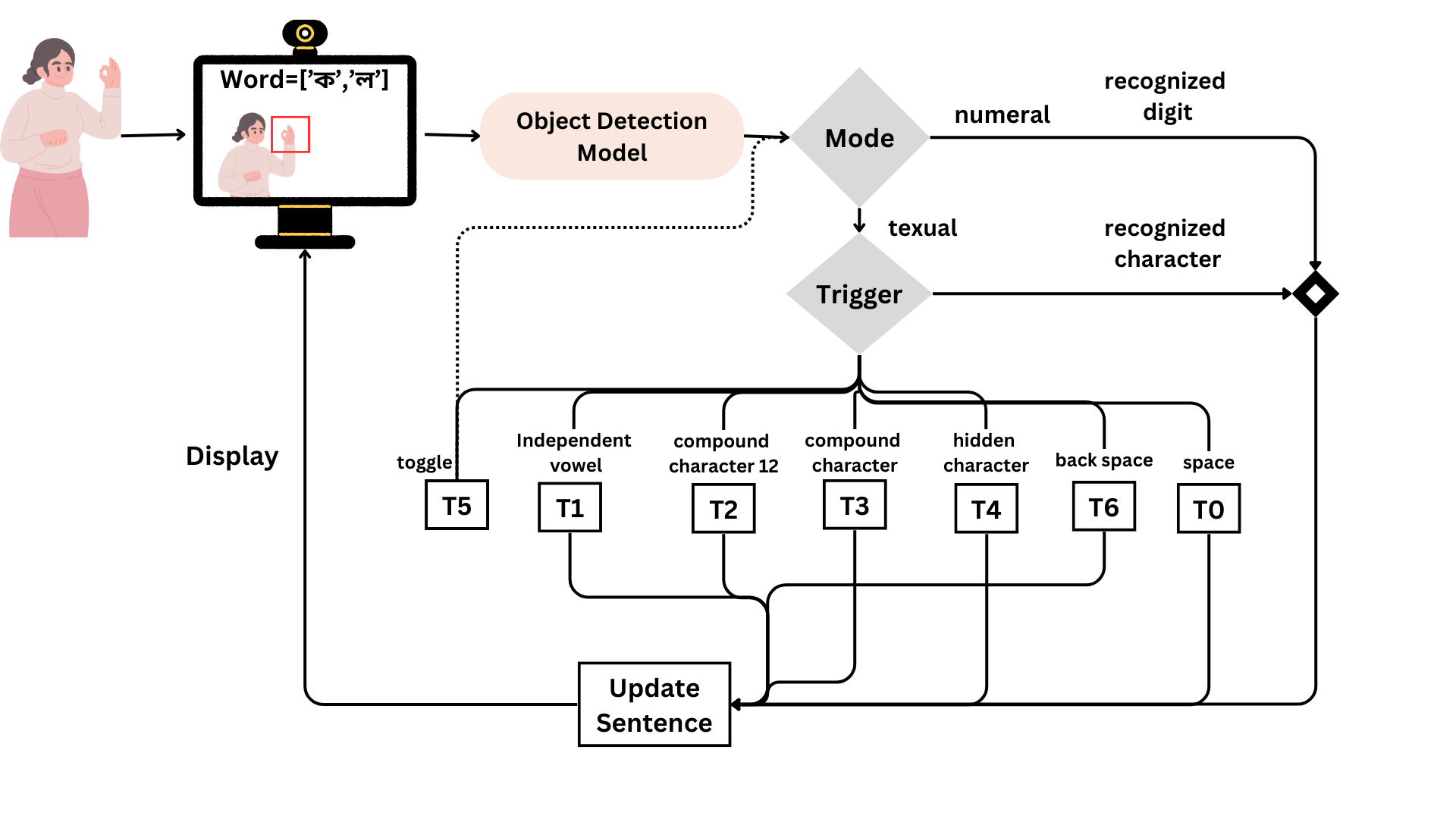}
\caption{This image depicts an algorithm utilizing YOLO v5 object detection to recognize characters. Recognized characters can transition into either numeral or textual modes. Textual mode is determined by triggers: T5 for toggles, T1 for independent vowels, T2 for compound character 12, T4 for hidden characters, T6 for backspace, and T0 for spaces.}\label{method}
\end{figure}

\subsubsection{Real Time Finger-Spelling}
The methodology for developing the BdSL Fingerspelling model is designed to accurately recognize finger-spelled Bengali characters in real time. The BDSL36 dataset, comprising 36 recognized characters and their derivatives, forms the basis of this system. Key components within the methodology include Recognized Character Detection, Independent Vowel Transformation, Hidden Character Generation, and Trigger Handling. Recognized characters are identified based on confidence scores generated by the YOLOv5 model, ensuring consistent high confidence for selection. Independent vowels are assumed to be dependent, allowing for accurate transcription, while hidden characters play a crucial role in representing Bengali script. Transformation rules and triggers facilitate the derivation of characters not present in the BDSL36 dataset. Compound characters are also derived using triggers and a Bangla compound character dictionary. The comprehensive approach of this methodology addresses the complexities of the Bengali language, resulting in a robust and versatile finger-spelling recognition system.

Our real-time finger spelling starts with a speller hand signing the recognized characters which in real-time contributes to the errors because of the ambient and other factors. In order to confirm a detection, we created a window that uses the confidence score, as explained in the above subsection. A detected character then passes through the trigger-handling module. Based on the Trigger, the recognized character is further transformed according to the flow chart shown in Figure \ref{method}.

In our system, we can finger-spell either texts or numerals, which can be done in the textual or numeral mode respective. By default, we begin with textual mode, the mode will only change if the trigger T5 (recognized character 5) is detected switching to a numeral mode where any numeral recognized character will not be detected as a trigger. To get back to the previous mode 'aa' has to be detected which acts as trigger T5 in the numeral mode. Each recognized character detected or transformed character is then added to update a sentence. These steps are shown in the Figure \ref{method}. Furthermore, we use trigger T0 to add space and trigger T6 to delete the last appended character. 

Using our setup and methodology explained above and in Figures [\ref{bengali_alphabets} \ref{compound_derivation}], we further demonstrate the finger spelling using examples in Figure [\ref{wfingerSpelling}].

\begin{figure}[h!]%
\centering
\includegraphics[width=\textwidth]{word_finger_spelling.png}
\caption{
In our fingerspelling system for Bengali, we break down the intricate art of conveying Bengali words using hand shapes and finger movements. Each word is meticulously spelled out using a combination of handshapes, finger motions, and specific positions, all in accordance with the unique script and phonetic characteristics of the Bengali language. This table offers a comprehensive guide to the fingerspelling of common Bengali words, making it an invaluable resource for those learning sign language or communicating with individuals who rely on this method to express themselves in Bengali.}\label{wfingerSpelling}
\end{figure}
\begin{figure}[h!]%
\centering
\includegraphics[width=\textwidth]{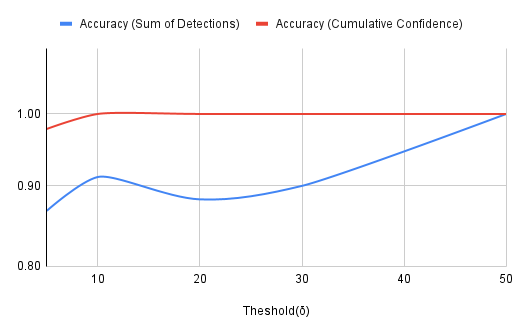}
\caption{In this graph, we explore the relationship between threshold values ($\delta$) and accuracy, considering two distinct measures: "Accuracy (Sum of Detections)" and "Accuracy (Cumulative Confidence)." As the threshold value varies along the X-axis, we observe changes in accuracy on the Y-axis. Lower threshold values (5 and 10) yield high accuracy levels in both measures, with "Accuracy (Cumulative Confidence)" reaching a perfect score at 10. Accuracy remains consistently high as the threshold increases, even reaching 100\% for "Accuracy (Cumulative Confidence)" at threshold values of 20, 30, and 50. This visual representation elucidates how threshold values impact detection accuracy, offering insights into system performance.}\label{detection_accuracy}
\end{figure}
\begin{figure}[h!]%
\centering
\includegraphics[width=\textwidth]{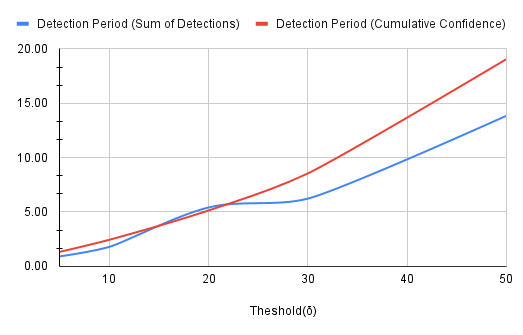}
\caption{In this graph, we explore how different threshold values ($\delta$) affect detection time. Lower thresholds (5 and 10) result in faster detections, especially in "Time (Cumulative Confidence)." However, as the threshold increases, both "Time (Sum of Detections)" and "Time (Cumulative Confidence)" take longer. Overall, this graph shows the balance between accuracy and detection speed based on threshold choices.}\label{detection_period}
\end{figure}
\section{Training YOLOv5}\label{sec4}
During the development of the project, the YOLOv5 model was trained on a dataset of Bangla Sign Language images to detect and classify different signs. After training, the model's performance was evaluated on a separate validation dataset to assess its accuracy and effectiveness.
The validation process used the best-trained weights of the model, which were saved at the location runs/train/exp/weights/best.pt. These weights represent the model's parameters that achieved the highest level of performance during the training process.
The model's architecture consists of 157 layers and is relatively lightweight, with 7,136,884 parameters. It's essential to have a model with a suitable number of parameters to strike a balance between accuracy and computational efficiency. In this case, the model's computational cost is measured in GFLOPs (Giga Floating Point Operations) and is found to be 16.2 GFLOPs, indicating a reasonable computational load.
To evaluate the model's detection performance, it was tested on a validation set comprising 1,826 images, which collectively contained 1,827 instances of Bangla Sign Language signs. The model's performance is measured using several metrics that provide insights into its ability to recognize different signs.

\section{Results}\label{sec5}
The primary evaluation metrics used are Precision (P), Recall (R), and Mean Average Precision (mAP) at various IoU (Intersection over Union) thresholds. Precision measures the accuracy of the model's predictions, reflecting the percentage of true positive detections among all the positive detections. Recall, on the other hand, measures the model's ability to find all the positive instances, indicating how well it avoids missing any relevant signs. The Mean Average Precision (mAP) is a comprehensive measure that considers the precision-recall trade-off across multiple IoU thresholds. It provides a holistic assessment of the model's performance for different classes and thresholds. In this evaluation, mAP is calculated at the standard IoU threshold of 0.5 and also across a range of IoU thresholds from 0.5 to 0.95.

The overall performance of the model across all classes combined is quite promising. It achieved a precision of 69.2\% and a recall of 83.6\%, indicating that it can identify a significant portion of Bangla Sign Language signs accurately. The mAP at the standard IoU threshold of 0.5 is 84.3\%, which is considered a strong performance. It means that the model's predictions are well-matched with the ground truth annotations.

However, the more comprehensive mAP across IoU thresholds from 0.5 to 0.95 is 56.9\%, which suggests that the model's performance may vary across different levels of detection strictness. It is common to see a drop in mAP as the IoU threshold increases, as it requires stricter overlap between predicted and ground truth bounding boxes. Looking into the performance of individual classes, we can observe variations in the model's ability to recognize different signs. Some classes, such as "BHA," "BISHARGA," and "THA," achieved high precision, recall, and mAP scores, indicating that the model performs exceptionally well on these signs.

On the other hand, there are classes like "NA" and "RA" with lower precision, recall, and mAP scores, suggesting that the model struggles more with recognizing these particular signs. In summary, the YOLOv5 model showed overall promising results for the Bangla Sign Language Detection project. It demonstrated the ability to detect and classify signs with good accuracy and achieved a high mAP score at the standard IoU threshold of 0.5. However, there are certain classes where the model's performance could be further enhanced. This might involve additional data collection for underrepresented classes, fine-tuning hyperparameters, or exploring other techniques to improve recognition accuracy. Continuous evaluation and refinement of the model will be essential to enhance its capabilities and make it more robust for real-world applications.

\begin{table}[h]
\centering
\label{comparision}
\begin{tabular}{|c|c|c|c|c|}
\hline
Model & Data & Dataset Size & Number of Classes & mAP \\
\hline
YOLOv5 & BdSL-OD & 1,000 images & 30 signs & 95\% \\
YOLOv4 \cite{wordgeneration} & BdSL-OD & 12,500 images & 49 signs & 95.6\% \\
YOLOv5 (ours) & BdSL 36 & 9,147 images & 36 signs & 96.40\% \\
\hline
\end{tabular}
\caption{Comparison of Object Detection Models}
\end{table}

In this comparison table \ref{comparision}, we present an overview of different object detection models used for Bangla Sign Language detection, highlighting key attributes such as the model name, dataset used, dataset size, number of classes, and mean Average Precision (mAP) scores. Notably, our YOLOv5 model achieved a superior mAP of 96.40\% on the BdSL 36 dataset comprising 9,147 images and 36 signs, outperforming the previous YOLOv4 model (95.6\%) from \cite{wordgeneration}. This improved performance may be attributed to architectural enhancements, a larger and diverse training dataset, the use of data augmentation techniques, fine-tuning for Bangla Sign Language, consistent evaluation metrics, and potential algorithmic advancements. Notably, YOLOv5 showcases better precision in sign detection, making it a promising advancement in the field, although potential limitations should also be acknowledged for a comprehensive evaluation of its performance.

\begin{figure}[h!]%
\centering
\includegraphics[width=1.2\textwidth]{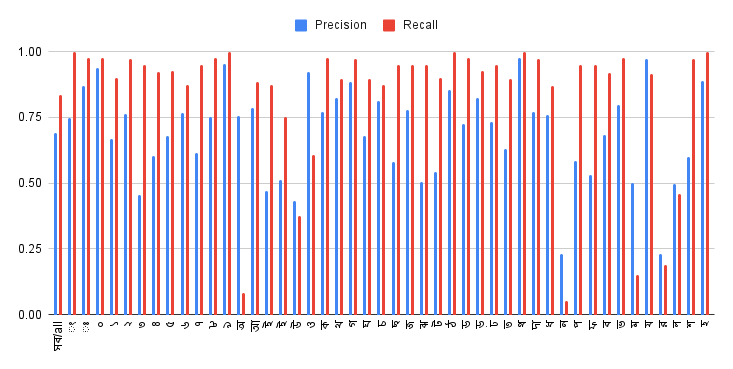}
\caption{Precision-Recall graph showcasing object detection performance by class, with mAP50 and mAP50-95 scores on the Y-axis. Evaluate model accuracy and variation across different characters.}\label{precision-recall-chart}
\end{figure}

The graph \ref{precision-recall-chart} displays the Precision-Recall performance for different classes, with the X-axis showing the classes, including both Bengali script and Romanized labels, and the Y-axis representing the mAP50 (mean Average Precision at 50\%) and mAP50-95 (mean Average Precision from 50\% to 95\% overlap) scores. This graph offers valuable insights into the model's object detection capabilities. For each class, the mAP50 score indicates how well the model can accurately identify instances of that specific character or symbol. Higher bars indicate better performance, while lower ones suggest areas where the model may struggle. The mAP50-95 scores provide a more comprehensive evaluation, considering a broader range of overlap thresholds. Overall, this graph allows us to assess the precision and recall performance of the model across different classes, identifying its strengths and weaknesses in object detection tasks.

\begin{figure}[h!]%
\centering
\includegraphics[width=0.9\textwidth]{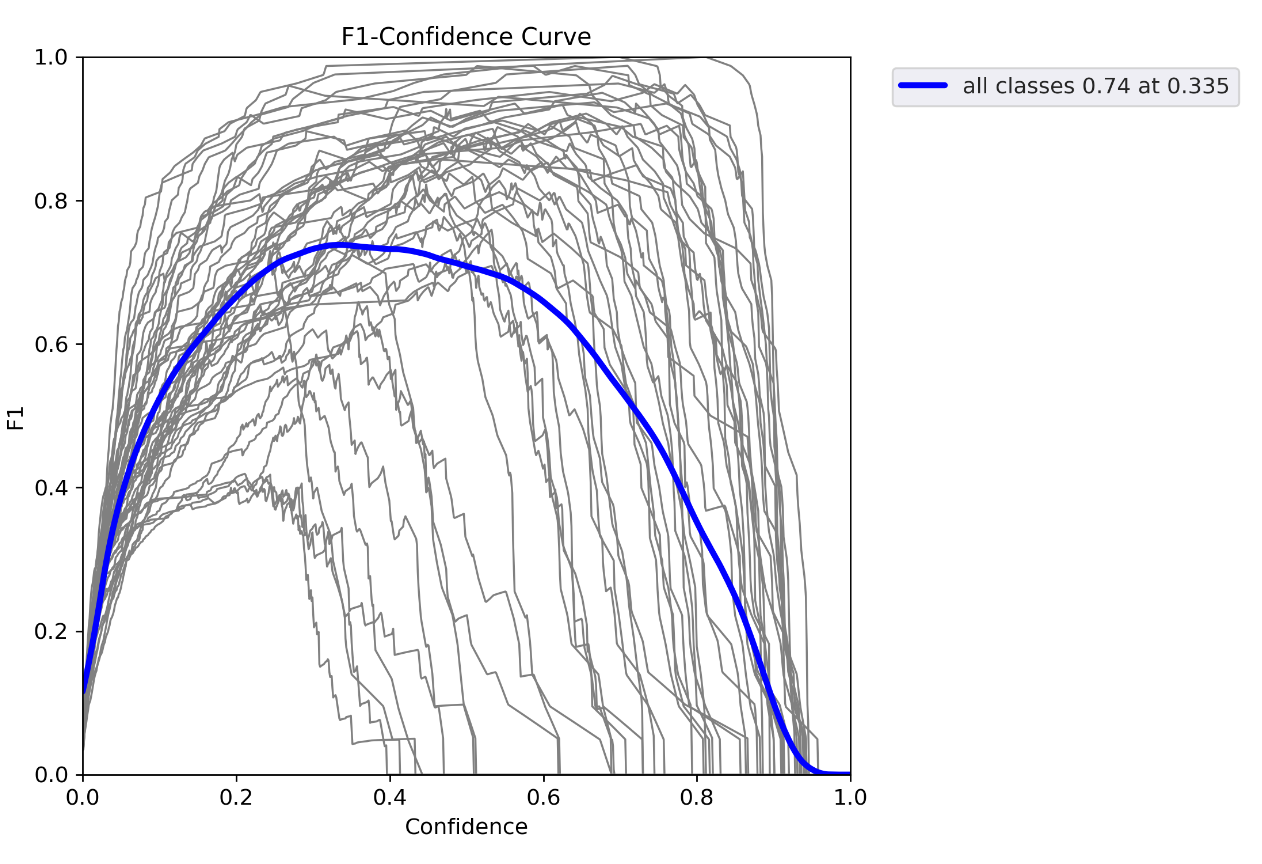}
\caption{This F1-confidence curve in the YOLO model demonstrates a promising performance for object detection. Achieving an F1 score of 0.74 at a confidence threshold of 0.335 indicates a good balance between precision and recall for detecting objects across all classes. The curve suggests that the model can identify objects accurately while minimizing false positives, making it suitable for object detection tasks.}\label{fig1}
\end{figure}

\begin{figure}[h!]\centering
\includegraphics[width=0.9\textwidth]{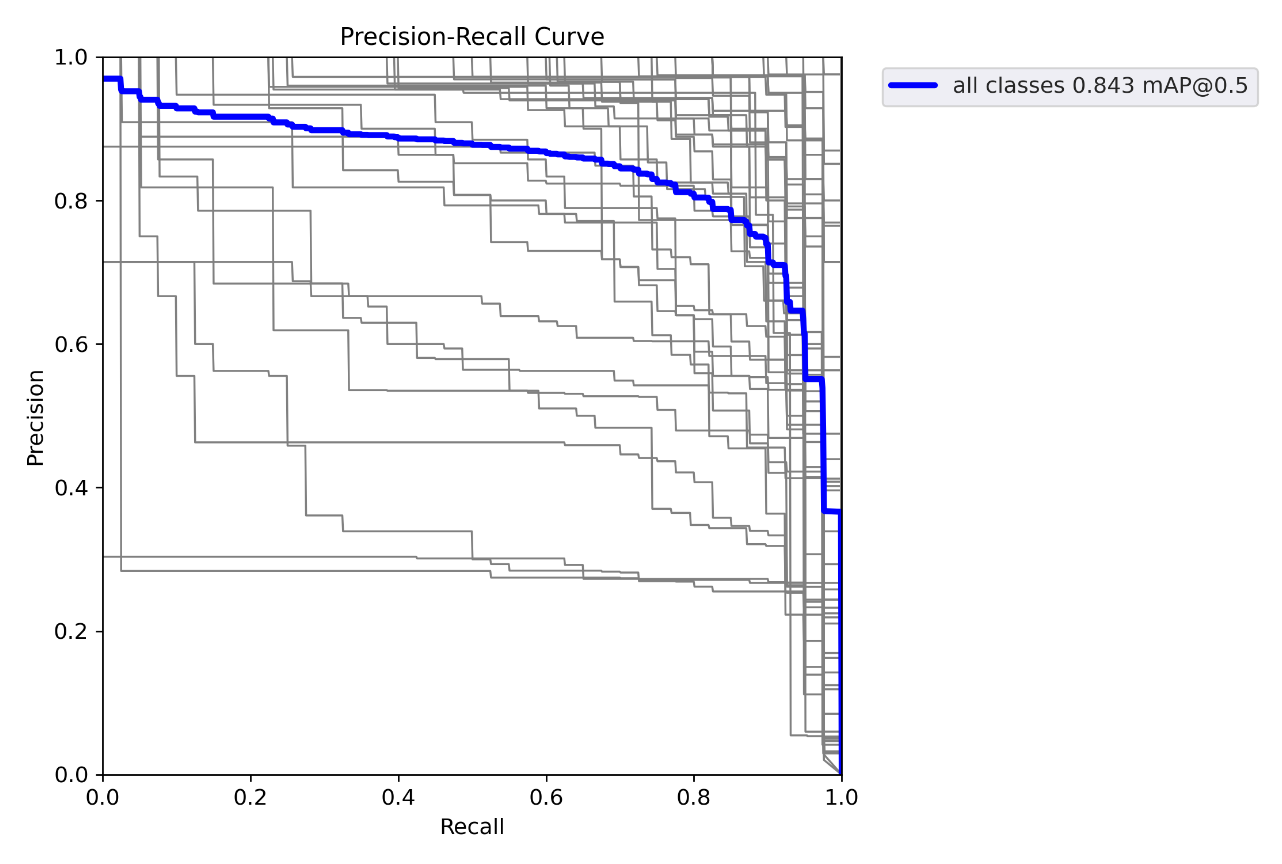}\caption{This precision-recall curve, achieving an mAP (mean Average Precision) of 0.74 at a confidence threshold of 0.5, indicates strong performance in object detection training with the YOLO model for all classes. An mAP of 0.74 signifies that the model can accurately identify objects with a high level of precision and recall, making it well-suited for various object detection tasks. The curve suggests that the model maintains a good balance between correctly identifying objects (precision) and capturing all relevant objects (recall) at this confidence threshold, demonstrating its effectiveness for object detection across different classes.}\label{fig1}
\end{figure}

\begin{figure}[h!]%
\centering
\includegraphics[width=0.9\textwidth]{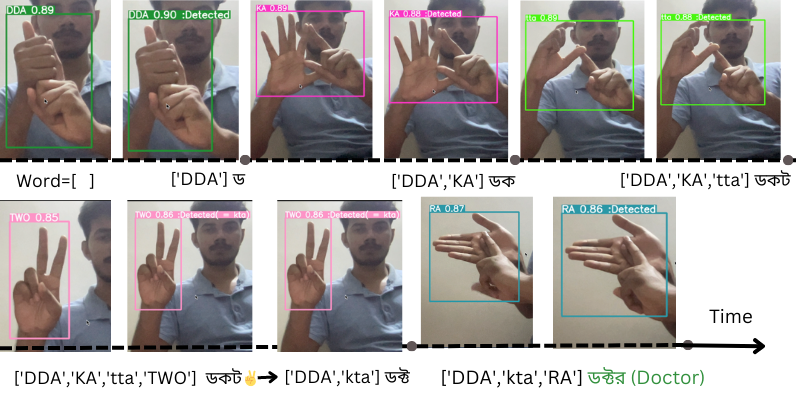}
\caption{This is a widefig. This is an example of long caption this is an example of long caption  this is an example of long caption this is an example of long caption}\label{fig1}
\end{figure}

\begin{table}[htbp]
\centering
\caption{Threshold, Accuracy ($\sum n$), Accuracy ($\sum c_{i}$), Frequency $n$, Frequency $c$}
\label{tab1}
\begin{tabular}{cccccc}
\toprule
Threshold ($\delta$) & Accuracy ($\sum n$) & Accuracy ($\sum c_{i}$) & Frequency $n$ & Frequency $c$ \\
\midrule
5 & 87\% & 98\% & 69.15 & 61.33 \\
10 & 91\% & 100\% & 65.81 & 60.00 \\
20 & 88\% & 100\% & 68.00 & 60.00 \\
30 & 90\% & 100\% & 66.67 & 60.00 \\
50 & 100\% & 100\% & 60.00 & 60.00\\
\bottomrule
\end{tabular}
\end{table}

\begin{figure}[h!]%
\centering
\includegraphics[width=0.9\textwidth]{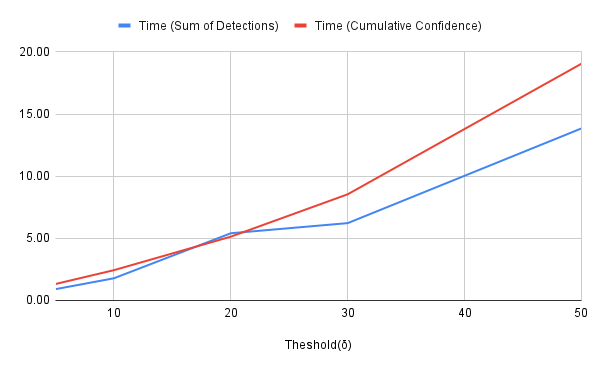}
\caption{}\label{fig1}
\end{figure}
\section{Conclusion}\label{sec13}
Conclusions may be used to restate your hypothesis or research question, restate your major findings, explain the relevance and the added value of your work, highlight any limitations of your study, and describe future directions for research and recommendations. 
In some disciplines use of Discussion or 'Conclusion' is interchangeable. It is not mandatory to use both. Please refer to Journal-level guidance for any specific requirements. 
\backmatter

\end{document}